\pdfoutput=1

\documentclass[11pt]{article}

\usepackage{emnlp2021}

\usepackage{times}
\usepackage{latexsym}
\usepackage{comment}
\usepackage{graphicx}
\usepackage{amsmath}
\usepackage{booktabs}
\usepackage{array}
\usepackage{microtype}
\usepackage{verbatim}
\usepackage{soul}
\usepackage{url}
\PassOptionsToPackage{hyphens}{url}
\usepackage{hyperref}

\usepackage[T1]{fontenc}

\usepackage[utf8]{inputenc}

\usepackage{microtype}

\newcounter{mycounter}
\newcommand\modelcounter{\stepcounter{mycounter}\textbf{(\themycounter)}}
\newcommand\labelledmodelcounter[1]{\refstepcounter{mycounter}\textbf{(\themycounter)}\label{model:#1}}
\newcommand{\modelref}[1]{\textbf{(\ref{model:#1})}}

\title{Multilingual Domain Adaptation for NMT: Decoupling Language and Domain Information with Adapters}

\author{Asa Cooper Stickland\thanks{~~Work done during an
    internship at NAVER LABS Europe.}  \\
  University of Edinburgh \\
  \texttt{a.cooper.stickland@ed.ac.uk} \\\And
  Alexandre B\'erard \qquad Vassilina Nikoulina \\
  NAVER LABS Europe \\
  \texttt{first.last@naverlabs.com}}

\begin{document}
\maketitle
\begin{abstract}
Adapter layers are lightweight, learnable units inserted between transformer layers. Recent work explores using such layers for neural machine translation (NMT), to adapt pre-trained models to new domains or language pairs, training only a small set of parameters for each new setting (language pair or domain).
In this work we study the compositionality of language and domain adapters in the context of Machine Translation. We aim to study, 1) parameter-efficient adaptation to multiple domains and languages simultaneously (full-resource scenario) and  2) cross-lingual transfer in domains where parallel data is unavailable for certain language pairs (partial-resource scenario).
We find that in the partial resource scenario a naive combination of domain-specific and language-specific adapters often results in `catastrophic forgetting' of the missing languages. We study other ways to combine the adapters to alleviate this issue and maximize cross-lingual transfer.
With our best adapter combinations, we obtain improvements of 3-4 BLEU on average for source languages that do not have in-domain data. For target languages without in-domain data, we achieve a similar improvement by combining adapters with back-translation. Supplementary material is available at \url{https://tinyurl.com/r66stbxj}.
\end{abstract}

\section{Introduction}

Multilingual Neural Machine Translation (NMT) has made a lot of progress recently \cite{johnson-etal-2017-googles, bapna-firat-2019-simple, aharoni-etal-2019-massively, zhang-etal-2020-improving, fan2020englishcentric} and is now widely adopted by the community and MT service providers. Multilingual NMT models handle multiple language directions at once and allow for knowledge transfer to low-resource languages. Machine translation systems often need to be adapted to specific domains like legal or medical text. However, when adapting multilingual systems, in-domain data for most language pairs might not exist. We would like to be able to leverage data in a subset of language pairs to transfer domain knowledge to other languages. 

Straightforward methods of domain adaptation include fine-tuning \cite{freitag2016fast} or usage of domain tags \cite{kobus-etal-2017-domain,britz-etal-2017-effective} for different domains. For these methods each new domain request would require re-training the whole model, which is a costly procedure. And naive training on a subset of languages typically reduces performance on all other languages \cite{forgetmulti}, a phenomenon known as `catastrophic forgetting' \cite{McCloskey1989CatastrophicII}.

An alternative technique for adapting such models to new language-pairs or domains are `adapter layers' \cite{bapna-firat-2019-simple}, lightweight, learnable units inserted between transformer layers. A previously trained large multilingual model can be adapted to each language-pair by learning only these small units, and keeping the rest of the model frozen. 
This procedure also allows for the incremental adding of new language pairs and/or domains to the pre-trained model, reducing the cost of adaptation. 
Previous studies have shown it is possible to combine language-specific (as opposed to language-pair specific) adapters \cite{philip-etal-2020-monolingual}, or language and task adapters \cite{pfeiffer-etal-2020-mad} trained independently, enabling zero-shot compositions of adapters. 
Our ultimate goal is, for ease of deployment and storage, a single model that can handle all languages and domains.
In this work we analyse how to combine \textit{language adapters} with \textit{domain adapters} in multilingual NMT, and study to what extent the  domain knowledge can be transferred across languages.

First, we show it is hard to  decouple language knowledge from domain knowledge when fine-tuning multilingual MT systems on new domains. In Section~\ref{sec:offtgt} we demonstrate that adapters learnt on a subset of language pairs fail to generate into languages not in that subset. Such generation into the wrong language is referred to as `off-target' translation. We additionally find combinations of domain and language adapters not seen at training time lead to bad performance.
We examine how adapter placement and other techniques can improve the compositionality of language and domain adapters when dealing with source or target languages that do not have in-domain data (which we refer to throughout this work as ``\textbf{out-of-domain languages}'').
Our key contributions are:
\begin{itemize}
    \item We examine domain adaptation capacity in the multi-lingual, multi-domain setting. We find that encoder-only adapters can be just as effective as default adapters added in every layer, and that composing domain adapters with language adapters outperforms language adapters alone, although fine-tuning with domain tags performs better for most domains. 
    \item We improve the cross-lingual transfer of domain knowledge for adapters. We analyse different language and domain adapter combinations that improve performance and reduce off-target translations. Our best results for translation into out-of-domain languages use decoder-only domain adapters, regularisation with domain adapter dropout, and data augmentation with English-centric back-translation. 
\end{itemize}

\section{Related Work}

\paragraph{Cross-lingual transfer}
Many works have demonstrated that large pre-trained multilingual  models \cite{devlin-etal-2019-bert, conneau-etal-2020-unsupervised,liu-etal-2020-multilingual} fine-tuned on high-resource languages (or language pairs) can transfer to lower-resource languages in various tasks: Natural Language Inference \cite{Conneau2018xnli}, Question Answering \cite{Clark2020tydiqa}, Named Entity Recognition \cite{pires-etal-2019-multilingual,K2020Cross-Lingual}, Neural Machine Translation \cite{liu-etal-2020-multilingual} and others \cite{hu2020xtreme}. 
\paragraph{Domain adaptation in NMT} Domain adaptation has been discussed extensively for  bilingual NMT models. A typical approach is to fine-tune a model trained on a large corpus of `generic' data on a smaller in-domain corpus \cite{Luong-Manning:iwslt15,neubig2018rapid}. A common technique to make use of monolingual in-domain data is to do back-translation \cite{sennrich-etal-2016-improving,berard-etal-2019-machine,jin2020simple}. Although effective, it is expensive to create back-translated data, especially when one needs to cover multiple language pairs. 
Multi-domain models can be trained with domain tags \cite{kobus-etal-2017-domain,britz-etal-2017-effective,berard-etal-2019-machine,stergiadis2021multidomain} that can encode domain-specific information. However, domain tags do not allow \textit{incrementally} adding new domains to a model: each new domain adaptation requires retraining the full model (as opposed to adapter layers that can be trained independently for each language/domain).
There are a number of works \cite{jiang-etal-2020-multi-domain, britz-etal-2017-effective, dabre2020comprehensive} trying to explicitly decouple domain-specific representations from domain independent representations in bilingual settings. In our work we try to decouple language and domain specific representations through adapter layers. 

\paragraph{Adapter layers} \citet{bapna-firat-2019-simple} introduce adapter layers for NMT as a lightweight alternative to fine-tuning. They study both adding language-pair specific adapters to multilingual NMT models to match the performance of a bilingual version, and domain-specific adapters for parameter-efficient domain adaptation. 
\citet{philip-etal-2020-monolingual} train adapters for each \textit{language} instead of \textit{language-pair} and show that composing such adapters 
improves zero-shot translation in English-centric settings, and can adapt a model to all language directions in a scalable way. \citet{pfeiffer-etal-2020-mad,pfeiffer-etal-2021-adapterfusion} study adapter layers for pre-trained language models evaluated on NLU tasks. They show it is possible to compose language and task adapters. Combining language adapters trained with a masked language modelling objective for language $x$ and task adapters trained on a classification task in language $y$ can transfer to classification in language $x$. 
We have a similar objective to \citet{pfeiffer-etal-2020-mad}, but for NMT where in addition to encoding sentences we need to generate text for new language and domain combinations. 

To the best of our knowledge none of the works above study composing language and domain adapters for generation tasks (such as translation) which is the goal of this work. 

\section{Composing Adapter Modules}
Adapter modules \cite{rebuffi2017learning,houlsby2019parameter} are randomly initialised modules inserted between the layers of a pre-trained network and fine-tuned on new data. An adapter layer is typically a down projection to a bottleneck dimension followed by an up projection to the initial dimension, which we write as $\textrm{FFN}(\textbf{h}) = W_{\textrm{up}}f(W_{\textrm{down}}\textbf{h})$, with $f(\cdot)$ a non-linearity.
The bottleneck controls the parameter count of the module; typically NMT requires slightly larger parameter counts than classification to match fine-tuning \cite{bapna-firat-2019-simple, cooper-stickland-etal-2021-recipes}. 
With a residual connection and a near-identity initialization the original model is (approximately) retained at the beginning of optimization, keeping at least the performance of the parent model.

\subsection{Stacking Domain and Language Adapters}
\label{sec:stack}
In this work we study `stacking' adapter modules, i.e. each language and domain has a unique adapter module associated with it. 
When passing a batch with source language $x$, target language $y$, and domain $z$, we only `activate' the adapters for $\{x, y, z\}$. 
The encoder adapters for $x$ and decoder adapters for $y$ are activated.

We mostly follow the architecture of \citet{bapna-firat-2019-simple}. Language adapters $\textrm{LA}$ are defined as:
\begin{equation}
    \textrm{LA}(\textbf{h}_l) = \textrm{FFN}_{\textrm{lg}}(\textrm{LN}_{\textrm{lg}}(\textbf{h}_l)) + \textbf{h}_l
\end{equation}
where $\textbf{h}_l$ is the Transformer hidden state at layer $l$ and $\textrm{LN}_{\textrm{lg}}$ is a newly initialised layer-norm. Let $\textbf{z} = \textrm{LA}(\textbf{h}_l)$; when stacking domain and language adapters, the layer output $\textbf{h}_{l,\textrm{out}}$ is given by:
\begin{equation}
    \textbf{h}_{l,\textrm{out}} = \textrm{FFN}_{\textrm{dom}}(\textrm{LN}_{\textrm{dom}}(\textbf{z})) + \textbf{z}
     \label{eq:bapna}
\end{equation}
For all models without any stacking we obtain layer output as in Eq.~\ref{eq:bapna} but replace $\textrm{LA}(\cdot)$ with the identity operation.

\citet{pfeiffer-etal-2020-mad} use a different formulation that empirically performed well for them, but that in initial experiments produced worse results in our setting. We list the corresponding equations and results in Appendix~\ref{sec:mad} and Appendix~\ref{sec:add}.
\subsection{Improving the Compositionality of Adapters}
\label{sec:improve}

In our initial experiments (Section \ref{sec:zsres}) we found that (unlike \citealp{pfeiffer-etal-2020-mad}) naive stacking of language and domain adapters does not work very well for unseen combinations of language and domains, and often results in off-target translation (i.e. translations into the wrong language).
Therefore, we study several strategies to improve the compositionality of adapters in the context of NMT:

    1) Using \textbf{decoder-only} domain adapters when translating from an out-of-domain source language into an in-domain\footnote{Reminder we refer to the subset of languages we have parallel data for in a particular domain as `in-domain', and all other languages as `out-of-domain'.} target language, and \textbf{encoder-only} domain adapters when translating from an in-domain source language into an out-of-domain target language. This means we never stack together a combination of language and domain adapter that was not seen at training time. We also find empirically that decoder-only adapters work well with back-translation, perhaps because they can `ignore' the noisy synthetic source-side data.
    
    2) \textbf{Domain adapter dropout} (DADrop). Similar to layer-drop \cite{Fan2020Reducing} but specialised to adapter layers, or AdapterDrop \cite{adrop} but without targeting specific layers, we randomly `drop' (i.e.\ skip) the domain adapter\footnote{We could additionally drop the language adapter, but since this was frozen in many experiments we limit ourselves to domain adapters for simplicity} and only pass the hidden state through the language adapter. This means the adapter stack in the layer above can more easily adapt to unfamiliar input, and encourages domain and language adapters to be more independent of each other. 
    
 3) \textbf{Data augmentation}. We often have access to monolingual data in a domain even when no parallel data is available. In this work we leverage English-centric back-translation (BT), i.e.\ translating monolingual data in some languages into English (thus avoiding the more expensive step of translating from each language into every other language). We examine the ability of such data to help cross-lingual transfer to unseen combinations of source and target language (BT means we have artificial data for every language in combination with English). We briefly explore `\textit{denoising auto-encoder}' style objectives as in unsupervised MT \cite{lample2017unsupervised} or sequence-to-sequence pre-training \cite{lewis2019bart}.

\section{Experimental Settings}
\subsection{Data}
For studying the domain transfer across languages we select four diverse domains that have data available in most language directions: translations of the Koran (\textbf{Koran}); medical text from the European Medicines Agency (\textbf{Medical}); translation of TED Talks transcriptions (\textbf{TED});  various technical IT text, e.g.\ the Ubuntu manual (\textbf{IT}).
All data was obtained from the OPUS repository \cite{TIEDEMANN12.463}. 
We create validation and test sets of around 2000 sentences each, and avoid overlap with training data (including parallel sentences in any language) with a procedure described in Appendix~\ref{sec:appendix}. Note that Medical, Koran and IT are from the same source as those of \citet{aharoni2020unsupervised}, although the train/test splits are different due to expanding the number of languages and wanting a consistent pipeline for obtaining the data.

\begin{figure*}
    \centering
    \includegraphics[width=0.72\linewidth]{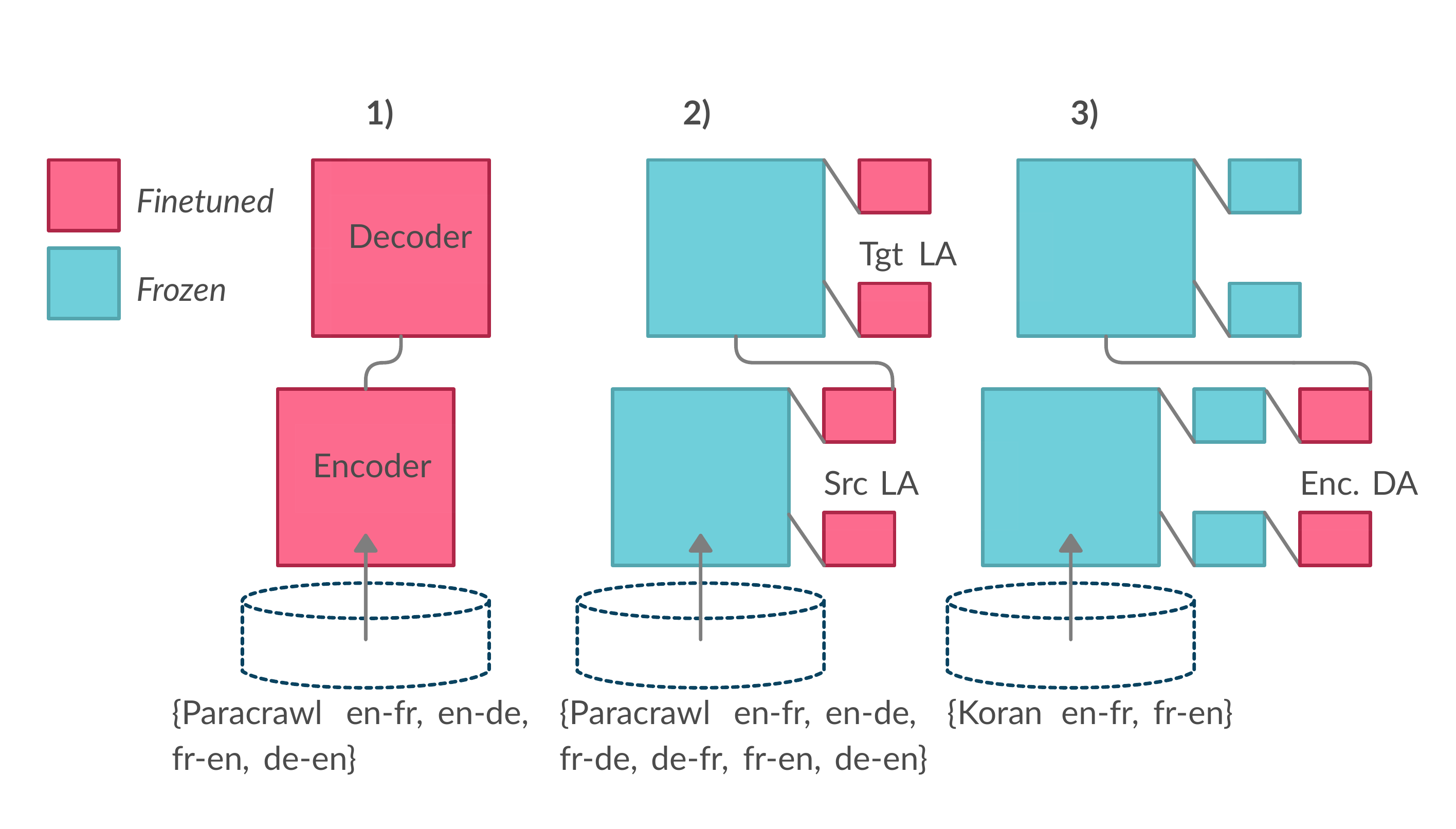}
    \caption{Toy diagram showing one of our proposed pipelines for training language and domain adapters, on a example subset of languages: $\{$en,fr,de$\}$, with `domain-agnostic' data from ParaCrawl and specialised data from the Koran. Red indicates a fine-tuned model component, blue indicates a frozen component. LA = language adapter, DA = domain adapter. From left to right we show: 1) Training an encoder-decoder model with English-centric ParaCrawl. 2) Training mononlingual language adapters with multiparallel Paracrawl data. 3) Training domain adapters stacked on language adapters in the encoder, on a subset (here $\{$en, fr$\}$) of languages for the domain of interest (e.g.\ Koran). Here we show domain adapters added only to the encoder, but we consider various other configurations in this work.}
    \label{figure:adapter-training-procedure}
\end{figure*}

\begin{table}[ht!]
    \centering
    \begin{tabular}{ccc}
    \toprule
        Domain & Langs. & Avg size (lines) \\
        \midrule
        ParaCrawl & 12 & 125M \\
        Koran & $10^{\dagger}$ & 52k \\
        Medical & $11^{\ddagger}$ & 500k \\
        IT & 12 & 196k \\
        TED & 12 & 138k \\
    \end{tabular}
    \caption{Basic statistics for the datasets we use; number of languages covered, and average number of training examples across all language directions. $\dagger$: missing nb \& da, $\ddagger$: missing nb.}
    \label{tab:data_stats}
\end{table}

\subsection{Models}

In \textbf{multilingual settings} we concentrate on 12 high-resource European languages\footnote{\{cs, da, de, en, es, fr, it, nb, nl, pl, pt, sv\}} due to the availability of domain-specific parallel data for most language pairs.
Our \textbf{baseline model} is a Transformer Base \cite{vaswani2017attention} trained on English-centric ParaCrawl v7.1 data \cite{banon-etal-2020-paracrawl} with all 12 languages (803M line pairs in total). It is trained with fairseq \cite{ott-etal-2019-fairseq} for 800k updates, with a batch size of maximum 4000 tokens and accumulated gradients over 64 steps \cite{ott-etal-2018-scaling}.\footnote{This corresponds to an effective batch size of $\approx$207k tokens and training length of 7 epochs.} The source/target embeddings are shared and tied with the output layer.
We tokenize the data with a shared BPE model of size 64k with inline casing \cite{berard2019}  Both the multilingual models and BPE model are trained with temperature-based sampling with $T=5$ \cite{mt-wild}. We calculate all BLEU scores with Sacrebleu\footnote{Signature: BLEU+case.mixed+lang.m2m-en+numrefs.1+smooth.exp+tok.13a+version.1.5.0.} \cite{post-2018-call}. On the recommendation of \citet{mtcred} we additionally report chrF \cite{popovic-2015-chrf} calculated using Sacrebleu\footnote{Signature: chrF2+numchars.6+space.false+version.1.5.1} for most models in the Appendix. We use adapter bottleneck size of 1024 unless stated otherwise, and when using DADrop (Section~\ref{sec:improve}) use a 20\% chance of skipping the domain adapter.

We additionally train monolingual language adapters \cite{philip-etal-2020-monolingual} for all 12 languages on multi-parallel ParaCrawl data, which we obtain by aligning all languages through their English side, like \citet{freitag-firat-2020-complete}. The adapters are trained for another 1M steps, without accumulated gradients.
We report the results of models fine-tuned on both all the domains simultaneously, or each domain separately, with access to in-domain data available for all the languages. Both serve as a potential upper bound for cross-lingual transfer.

We train the same model (i.e.\ with access to all languages) with domain tags: one special token per domain prepended to each source sequence \cite{kobus-etal-2017-domain}. We also measure the cross-lingual transfer ability of domain tags, by training a model with domain tags on all 4 domains but with in-domain data in only 4 languages (fr, de, cs and en). Because the latter model exhibits catastrophic forgetting issues in the other languages, we also train the same model with ParaCrawl data in all language directions (with a ``paracrawl'' domain tag). ParaCrawl line pairs are sampled with probability $0.5$.
More training hyper-parameters are given in Appendix~\ref{sec:appendix}.

\subsection{Our model pipelines}
We perform two series of experiments. 

\textbf{Multilingual multi-domain models.} Firstly, we experiment with different ways of multi-domain adaptation of multilingual models. We adapt the English-centric ParaCrawl pre-trained model to four domains (Koran, Medical, IT and TED) and every language direction simultaneously. We test models with language adapters, language + domain adapters, and domain tags. There is no cross-lingual domain transfer needed\footnote{There is obviously cross-lingual domain transfer that may take place when all the domains are trained jointly, but we do not explicitly study this in the first experiment.} since all language directions are included in the training data. Results for this scenario are reported in Section~\ref{sec:mmd}.

\textbf{Cross-lingual domain transfer.}  In the second experiment we try to decouple the notion of domain from language via analysing the zero-shot composition of domain and language adapters. This is described in a toy diagram in Figure~\ref{figure:adapter-training-procedure}. We first extend the baseline multilingual English-centric model with 12 (one for each language) monolingual language adapters \cite{philip-etal-2020-monolingual} trained on multi-parallel ParaCrawl. We then test the cross-lingual domain transfer ability of our proposed combinations of adapters by training on data in a particular domain with a subset of four languages (referred to as `\textbf{in-domain}'; in Figure~\ref{figure:adapter-training-procedure} \textit{en} and \textit{fr} would be in-domain). We test our model on all language directions from the set of all twelve languages. This will include cases where we don't have in-domain data for either the source or target language, which we refer to as `\textbf{out-of-domain}' (in Figure~\ref{figure:adapter-training-procedure} \textit{de} would be out-of-domain).

Finally, we extend the above mentioned scenario with back-translated (BT) data from \textit{out-of-domain} languages into English. To create the BT data, we use the model with language adapters trained on ParaCrawl \modelref{ParaCrawlLA} (which has not seen any in-domain data) on the English-aligned training data for each language and domain, and use beam search with a beam size of 5. Results for this scenario are reported in Section~\ref{sec:offtgt}.

To train language and domain adapters, we freeze all model parameters except for adapter parameters, and use a fixed learning rate schedule with learning rate $5\times{}10^{-5}$. Following \citet{philip-etal-2020-monolingual}, when training language adapters without domain adapters we build homogeneous batches (i.e.\ only containing sentences for one language direction) and activate only the corresponding adapters. When training language \textit{and} domain adapters together, we build homogeneous batches that only contain sentences for the same combination of language direction and domain.

\begin{table*}[ht!]
\centering
\begin{tabular}{clccccc}
\toprule
ID & Model & IT & Koran & Medical & TED & Params (M)\\
\midrule
\labelledmodelcounter{base} & Base (En-centric) & 23.2 & 7.0 & 25.7 & 19.0 & N/A \\
\midrule
\labelledmodelcounter{finetune_all_domains} & Finetuned & 40.8 & 16.0 & 42.7 & 26.6 & 79 \\
\labelledmodelcounter{domain_tags_all} & Finetuned + domain tags & \textbf{43.6} & \textbf{20.3} & \textbf{46.0} & 27.2 & 79 \\
\midrule
\labelledmodelcounter{singleLA} & Single adapter per layer ($d=1024$) & 39.6 & 14.7 & 41.8 & 26.2 & 12.6 \\
\labelledmodelcounter{multiLA} & LA ($d=1365$) & 42.0 & 17.6 & 43.7 & 26.8 & 202 \\
\labelledmodelcounter{multiLA_2048} & LA ($d=2048$) & 42.2 & 18.1 & 43.8 & 26.9 & 303  \\
\labelledmodelcounter{multiLA_decDA}  & LA + dec. DA ($d=1024$) & 42.1 & 18.5 & 43.6 & 27.1 & 177 \\
\labelledmodelcounter{multiLA_encDA} & LA + enc. DA ($d=1024$) & 42.3 & 19.3 & 43.8 & 27.5 & 177 \\
\labelledmodelcounter{multiLA_DA} & LA + enc \& dec. DA ($d=1024$) & 42.7 & 20.1 & 44.0 & \textbf{27.7} & 202\\
\end{tabular} 
\caption{\label{tab:multi}
BLEU scores averaged across all the language-directions for various multilingual multi-domain adaptation strategies, i.e. training on all language directions from the 12 languages and all domains. LA = language adapters, DA = domain adapters. `Params (M)' refers to the number of trainable parameters in millions. Note that unlike in Table~\ref{tab:frdecs-koran} the LA here are not pre-trained on ParaCrawl; they are trained jointly with domain adapters.}
\end{table*}

\begin{table*}[ht!]
\centering
\begin{tabular}{clccccc}
\toprule
ID & Model & All & In$\to$in & Out$\to$in & In$\to$out& Out$\to$out \\
\midrule
& \multicolumn{1}{c}{\textit{Oracles}} \\ \hline
\labelledmodelcounter{finetune_all_langs} & Finetune (all langs) & 44.3 & 43.9 & 44.8 & 44.2 & 44.2 \\
\modelref{domain_tags_all} & FT (all langs \& domains) + dom. tags & 46.0 & 45.3 & 46.3 & 45.9 & 46.0 \\
\midrule
& \multicolumn{1}{c}{\textit{Baselines}} \\ \hline
\modelref{base} & Base (En-centric) & 25.7 & 27.0 & 27.2 & 25.9 & 24.3 \\
\labelledmodelcounter{ParaCrawlLA}  & \modelref{base} + ParaCrawl LA & 30.2 & 29.6 & 30.8 & 30.0 & 30.0 \\
\midrule
& \multicolumn{1}{c}{\textit{Straightforward Methods}}\\ \hline
\labelledmodelcounter{vanillaDA} & \modelref{base} +  Domain adapters only & 23.0 & 44.7 & 37.8 & 13.4 (7\%) & 13.4 (11\%) \\
\labelledmodelcounter{EncDecDA} & Freeze LA + enc. \& dec. DA & 26.9 & 44.0 & 36.7 & 20.1 (71\%) & 19.9 (76\%) \\
\labelledmodelcounter{freezeLA_encDA} & Freeze LA + enc. DA & 29.6 & 42.6 & 34.0 & 27.0 (89\%) & 24.6 (88\%) \\
\labelledmodelcounter{freezeLA_decDA} & Freeze LA + dec. DA & 29.0 & 41.7 & \bf{40.7} & 22.5 (77\%) & 22.0 (77\%) \\

\labelledmodelcounter{TagsCross_no_paracrawl} & FT (all domains) + dom. tags & 15.6 & \bf{46.8} & 13.2 (55\%) & 12.0 (1\%) & 10.7 (2\%) \\
\midrule
& \multicolumn{1}{c}{\textit{Improving Off-target Translation}}\\ \hline

\labelledmodelcounter{TagsCross} & \modelref{TagsCross_no_paracrawl} + ParaCrawl & 34.7 & 42.2 & 39.6 & 32.4 & 31.0 \\
\labelledmodelcounter{EncDecDA_BT} & \modelref{EncDecDA} + BT & 33.9 & 43.2 & 36.8 & 35.9 & 28.0 (85\%) \\
\labelledmodelcounter{freezeLA_encDA_BT} & \modelref{freezeLA_encDA} + BT & 32.5 & 41.8 & 35.0 & 34.7 & 26.8 (83\%) \\
\labelledmodelcounter{freezeLA_decDA_BT} & \modelref{freezeLA_decDA} + BT & \bf{36.9} & 40.9 & 38.2 & 36.4 & \bf{35.1} \\

\midrule
\labelledmodelcounter{EncDecDA_DADrop} & \modelref{EncDecDA} + DADrop & 28.0 & 42.6 & 36.7 & 22.9 (82\%) & 21.5 (82\%) \\

\labelledmodelcounter{EncDecDA_DADrop_BT} & \modelref{EncDecDA} + BT + DADrop & 34.8 & 42.2 & 37.0 & 36.5 & 30.2 \\
\midrule
\labelledmodelcounter{unfreezeLA_decDA} & Unfreeze LA + dec. DA & 14.3 & 45.8 & 36.6 & 0.0 (1\%) & 0.0 (2\%) \\
\labelledmodelcounter{unfreezeLA_decDA_DADrop} & \modelref{unfreezeLA_decDA} + DADrop & 31.1 & 45.5 & 36.9 & 23.9 (82\%) & 27.8 \\
\labelledmodelcounter{unfreezeLA_decDA_DADrop_BT} & \modelref{unfreezeLA_decDA} + DADrop + BT & 35.2 & 44.5 & 33.4 & \bf{38.2} & 31.8 \\
\end{tabular}
\caption{\label{tab:frdecs-koran}
BLEU score of various models trained on the $\{$en, fr, de, cs$\}$ subset of the Medical domain, except `Oracle' models trained on all language pairs. LA = language adapters, DA = domain adapters. `Out$\to$in' is the average score when translating from an out-of-domain source language into $\{$en, fr, de, cs$\}$. `In$\to$out' corresponds to when the out-of-domain language is the target language. `In$\to$in' refers to average score when source and target are in the set $\{$en, fr, de, cs$\}$. `Out$\to$out' is the average score when both the source and target  language are unseen during domain adaptation. We note percentage of \textbf{on-target} (correct language) translations in brackets, when it is less than 90\% only.
}
\end{table*}
\section{Results and Discussion}
\label{sec:res}
First, in Section~\ref{sec:mmd} we discuss the results of experiments testing the domain adaptation capacity of various models, assuming access to data for all language pairs.
In Section~\ref{sec:offtgt} we analyse domain transfer across languages with adapters and other methods. We first demonstrate problems with cross-lingual generalisation during domain adaption for `naive' methods, and then propose potential solutions. Note we concentrate on the medical domain and a particular language subset for convenience. Appendix~\ref{sec:add} has results in other domains and language subsets, and also chrF \cite{popovic-2015-chrf} scores; we find similar trends to those reported in Section~\ref{sec:offtgt}.

\subsection{Multilingual multi-domain models}
\label{sec:mmd}

Table~\ref{tab:multi} reports the results from the challenging task of adapting a multilingual NMT model to multiple domains and language directions simultaneously. In this scenario, we assume access to in-domain data in all the language directions, and so we are testing the capacity of various models for domain adaptation, rather than cross-lingual transfer. Models are compared against a baseline \modelref{base} not trained on in-domain data.

We report the results for naive fine-tuning on the concatenation of in-domain parallel datasets for all the languages and all the domains \modelref{finetune_all_domains}. On all domains we improve on these results by fine-tuning with domain tags \modelref{domain_tags_all} (a similar result to \citet{jiang-etal-2020-multi-domain} in the bilingual setting).

Fine-tuning with domain tags \modelref{domain_tags_all} outperforms the model with stacked adapters \modelref{multiLA_DA}. A fine-grained comparison of these models is in Figure~\ref{fig:mmd_st} in the Appendix. For the IT and Medical domains the model with tags \modelref{domain_tags_all} is clearly better for all language directions. For the lowest resource domains, Koran and TED, most of the differences are not statistically significant, except for English-centric language pairs for TED, where the adapter model \modelref{multiLA_DA} is better. Exploring the combination of domain tags and adapters could be an interesting future research direction.

Stacking domain and language adapters \modelref{multiLA_DA} results in better performance than a model with the same parameter budget devoted to language adapters only \modelref{multiLA}. We believe this is because it allows the model to (partially) decouple domain from language-specific information, and better exploit the allocated parameter budget. Even a higher capacity language adapter model \modelref{multiLA_2048} does not perform as well. 

We also note that usage of encoder-only domain adapters \modelref{multiLA_encDA} outperforms the decoder-only domain adapter model \modelref{multiLA_decDA}. This is perhaps because the encoder representations influence the whole model (it is directly connected to the decoder at all layers with encoder-decoder attention) as opposed to the the decoder adapters that only impact decoder representations. We find a similar trend in bilingual domain adaptation, see Appendix~\ref{sec:bil2}.

The strong performance of encoder-only adapters has interesting implications for inference speed. With an auto-regressive decoder, the computational bottleneck is on the decoder side. The encoder output is computed all at once, while computing the decoder output requires $L$ steps, where $L$ is the output length. This implies devoting more capacity to encoder adapters would achieve similar performance and faster inference (more details in Appendix \ref{sec:bil2}).

\subsection{Cross-lingual Domain Transfer}

To study the capacity of our models to transfer domain knowledge across languages, we perform domain adaptation using parallel datasets for a \textit{subset} of language pairs, and evaluate on the test sets available for \textit{all} language pairs. In this section we report the results for adaptation to the \textit{medical} domain using the subset of all the language directions including $\{$en, fr, de, cs$\}$ languages (Table~\ref{tab:frdecs-koran}). We refer to these languages as \textit{in-domain} languages, and \textit{out-of-domain} languages would include all the other languages, $\{$de, nl, sv, es, it, pt, pl $\}$ (referred to as \textit{In }and \textit{Out} respectively in Table~\ref{tab:frdecs-koran}).

We report BLEU scores averaged across  test sets of different categories of language-directions depending on whether the source/target language was observed during the domain adaptation training: In$\to$in for language pairs observed during DA, Out$\to$out for fully zero-shot DA performance, and In$\to$out, Out$\to$in for translation directions combining \textit{in-domain} and \textit{out-of-domain} languages.

First, we report the results for \textit{Oracle} models providing an upper bound for the scores models could achieve with access to in-domain data for \textit{all} the languages: model \modelref{finetune_all_langs} was fine-tuned on \textit{medical} data for all the language directions\footnote{This is different from the model \modelref{domain_tags_all} which was fine-tuned on \textit{all} the domains and all the language directions.}, and a model with domain tags \modelref{domain_tags_all} discussed in section~\ref{sec:mmd}. 

Baseline models include the default multilingual English-centric model \modelref{base}, as well as model \modelref{ParaCrawlLA} with language adapters trained on multi-parallel ParaCrawl data. Comparing against this baseline shows us improvements from domain-specific (rather than language-specific) information.

\paragraph{Straightforward Methods}
\label{sec:offtgt}

We train several `straightforward' adapter models for the subset of \textit{in-domain} languages on the top of the baseline model, one with no language adapters, model \modelref{vanillaDA}, and model \modelref{EncDecDA} with domain \textit{and} language adapters (where language adapters are frozen), stacking them in the encoder and decoder.

Both of these models achieve good scores when translating into \textit{in-domain} languages (the In$\to$in and Out$\to$in categories), on par or better than \textit{Oracle} scores and much higher than the baselines. On the other hand they suffer from significant drops in performance when translating into \textit{out-of-domain} languages (the In$\to$out and Out$\to$out categories). 

The model \modelref{TagsCross_no_paracrawl} trained with tags on a subset of \textit{in-domain} languages suffers from the same low performance translating into \textit{out-of-domain} languages and additionally has low performance with out-of-domain \textit{source} languages. 

Looking closer at the translations of the above models, we see that many translations are either generated in English, copy the source language, or mix words between English and the true target language; see Table~\ref{tab:ex_offtarget} in the Appendix for illustrative examples. We refer to this phenomenon as "off-target" translation. We report the percentage of translations generated in the correct target language in Table~\ref{tab:frdecs-koran} when it is lower than 90\% \footnote{This percentage is computed against the reference translations that were correctly tagged by `langdetect', a Python language identifier (\texttt{https://pypi.org/project/langdetect/}). This is to exclude very short and numerical examples which can be quite frequent in some domains.}.  

We believe this phenomenon is partly due to decoder domain adapters having never been exposed to \textit{out-of-domain} language generation. Encoder domain adapters seem to be less sensitive to composition with new language adapters (as observed by \citet{pfeiffer-etal-2020-mad} for NLU tasks, and Table~\ref{tab:frdecs-koran} in the Out$\to$In column). 

\begin{figure*}[t]
\begin{center}
    \includegraphics[width=\linewidth]{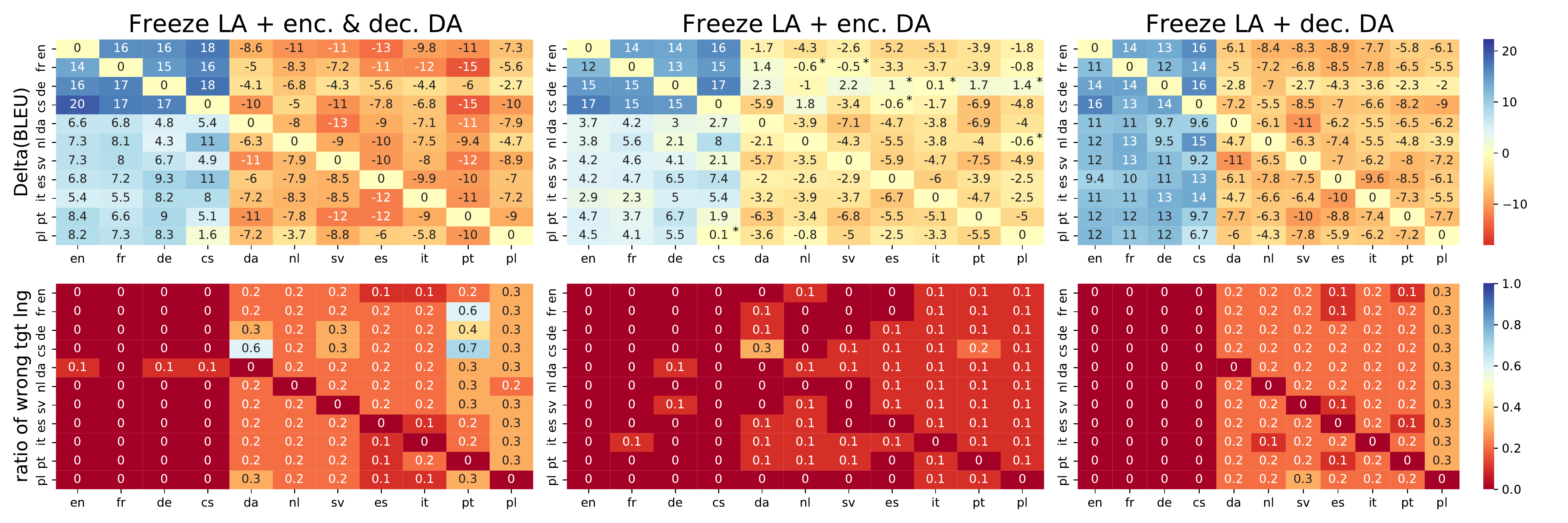}
\end{center}
    
   \caption{Comparing models with encoder-decoder adapters, encoder-only adapters and decoder-only adapters. $x$-axis shows the target language and $y$-axis shows the source language. Languages are grouped so the in-domain languages are in the top left corner. Top: Difference in BLEU compared to the baseline \modelref{ParaCrawlLA} (negative scores indicate a decrease w.r.t.\ the baseline, "*" indicates \textit{not} statistically significant). Bottom: proportion translating into the wrong target language. Best viewed in .pdf form.}
    \label{figure:adapter_place}
\end{figure*}

To investigate this, we train models \modelref{freezeLA_encDA} and \modelref{freezeLA_decDA} with encoder-only and decoder-only adapters. Figure \ref{figure:adapter_place} compares the performances of these models as well as model \modelref{EncDecDA} trained with encoder and decoder domain adapters, \modelref{freezeLA_encDA}, \modelref{freezeLA_decDA} against the baseline model \modelref{ParaCrawlLA}. The decoder-only model \modelref{freezeLA_decDA} can better translate \textit{from} out-of-domain languages and the encoder-only model \modelref{freezeLA_encDA} slightly improves for translations \textit{into} out-of-domain languages. However the problem of off-target translation persists for both models
and neither improves over ParaCrawl LA \modelref{ParaCrawlLA}. 
Therefore, we conclude that a straightforward combination of domain and language adapters leads to catastrophic forgetting both in the encoder and the decoder, but the encoder is less important for this effect. 

\paragraph{Effect of data augmentation}
\label{sec:zsres}

We train models \modelref{TagsCross},\modelref{EncDecDA_BT} \modelref{freezeLA_encDA_BT}, \modelref{freezeLA_decDA_BT} with additional data (either a portion of ParaCrawl data, or back-translation of in-domain data) to alleviate potential forgetting of representations for \textit{out-of-domain} languages. All of these models improve the translation quality into \textit{out-of-domain} languages. The model with tags \modelref{TagsCross} reaches competitive results and can be considered as a strong baseline.

\begin{figure*}
    \centering
    \includegraphics[width=\linewidth]{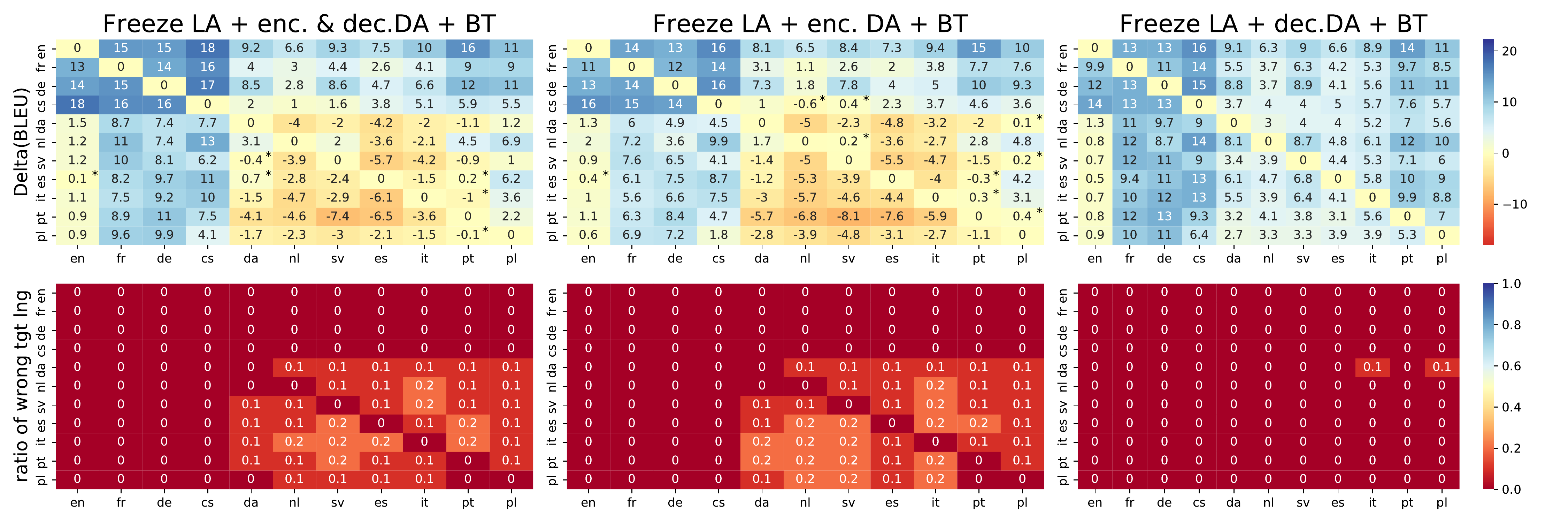}
    \caption{Comparing adapter models trained with back-translation. Top: Difference in BLEU compared to the baseline \modelref{ParaCrawlLA} ("*" indicates \textit{not} statistically significant). Bottom: proportion translating into the wrong target language. See Figure~\ref{figure:adapter_place} for more details.}
    \label{fig:adapter_place_bt}
\end{figure*}

For models with back-translation data, the decoder-only adapter \modelref{freezeLA_decDA_BT} model outperforms the encoder-only adapter \modelref{freezeLA_encDA_BT} model on out-of-domain target languages (as opposed to the case without BT) and has the strongest results overall on translating into out-of-domain languages. 
While the BT models are trained on exactly the same data, this effect is possibly due the encoder adapters being more influenced by potentially noisy synthetic source-side data, whereas decoder adapters are more influenced by clean reference translations. 
The decoder-only BT model \modelref{freezeLA_decDA_BT} improves over the baseline for all the language directions except for translation into English; see Figure \ref{fig:adapter_place_bt}.  

We report results for the other data augmentation methods (see Section~\ref{sec:improve}) in Appendix~\ref{sec:add}; these only improve over the ParaCrawl LA baseline in limited settings.

\begin{figure*}
\centering
\includegraphics[width=\linewidth]{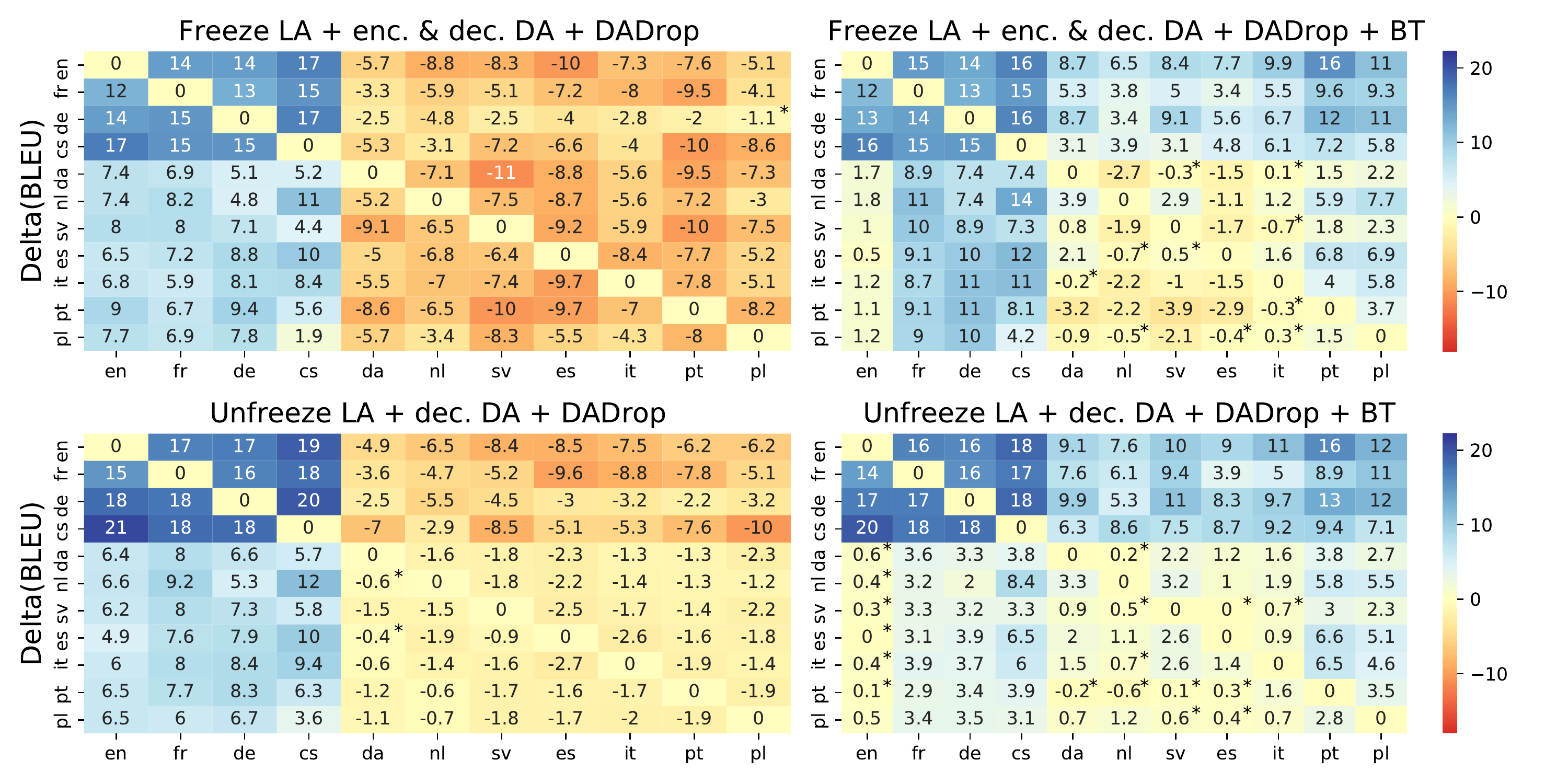}
\caption{Comparing models with DADrop, back-translation and unfrozen language adapters. Difference in BLEU compared to the baseline \modelref{ParaCrawlLA} ("*" indicates \textit{not} statistically significant). See Figure~\ref{figure:adapter_place} for more details.}
    \label{figure:bt_dropout}
\end{figure*}

\paragraph{Domain adapter dropout} Models  \modelref{EncDecDA_DADrop} and \modelref{EncDecDA_DADrop_BT} trained with dropping domain adapters (DADrop; see Section~\ref{sec:improve}) also allow to reduce catastrophic forgetting, although only combining DADrop with Data Augmentation (model \modelref{EncDecDA_DADrop_BT}) allows to solve the problem of off-target translation. 
We also note slight decreases in \textit{in-domain} performance for those models, perhaps due to underfitting. 

\paragraph{Increasing adaptation capacity} 
When naively increasing model capacity by unfreezing LA stacked with decoder DA  \modelref{unfreezeLA_decDA}, the model seems to mostly devote this capacity to In$\to$in category, and suffers on other language pair groups. This trend seems to be similar to the un-augmented model with tags \modelref{TagsCross_no_paracrawl}. However, once regularized with DADrop \modelref{unfreezeLA_decDA_DADrop}, and augmented with back-translation \modelref{unfreezeLA_decDA_DADrop_BT} it reaches very competitive results.

Figure \ref{figure:bt_dropout} shows fine-grained results for different models with DADrop, back-translation and unfrozen LA. 
Back-translation improves performance on the Out$\to$out and In$\to$out groups, but decreases performance on the Out$\to$in group. Finally, unfreezing language adapters decreases the performance on Out$\to$in but improves on the Out$\to$out group.

\paragraph{Adapters vs.\ tags}
As mentioned previously, model \modelref{TagsCross} with tags augmenteed with ParaCrawl reaches competitive scores overall. Note that this model was trained on a concatenation of \textit{all} the domains, unlike the models with adapters which were trained \textit{only} on the medical domain. Therefore it has been exposed to more data overall. 
On the other hand, several of our models fine-tune only a single adapter per-layer and use frozen LA. Thus, encoder-only or decoder-only models only require 6.3 million tunable parameters, compared to 79 million for tag-based models. Additionally adapter models can easily be `mixed-and-matched' by activating a particular adapter for a particular language pair. For example we could activate model \modelref{freezeLA_decDA} on `Out$\to$in' (out-of-domain source, in-domain target) data, model \modelref{EncDecDA_BT} on in-domain data and model \modelref{freezeLA_decDA_BT} otherwise. Such models could easily be extended to new domains by training more adapters, in contrast to tag-based models which update all parameters for each domain adaptation request.

\section{Conclusion}   
In this work we studied multilingual domain adaptation both in the full resource setting where in-domain parallel data is available for all the language pairs, as well as the partial resource setting, where in-domain data is only available for a small set of languages. 

In particular, we study how to better compose language and domain adapter modules in the context of NMT. We find that while adapters for encoder architectures like BERT can be safely composed, this is not true for NMT adapters: domain adapters learnt in the partial resource scenario struggle to generate into languages they were not trained on, even though the original model they are inserted in was trained on those languages. 
We found that randomly dropping the domain adapter and back-translation can regularize the training and lead to less catastrophic forgetting for when generating into out-of-domain languages, although they do not fully solve the problem of off-target translation. 

We experimented with different adapter placement and found that devoting additional capacity to encoder adapters can lead to better results compared to when the same capacity is shared between the encoder and the decoder. Similarly, in the partial resource scenario, models with encoder-only domain adapters suffer less from catastrophic forgetting when translating into out-of-domain languages. In contrast, decoder-only domain adapters perform well when translating \textit{from} out-of-domain into in-domain languages, and combine well with back-translation, perhaps due to their ability to ignore noisy synthetic source data.

Finally we note that a model fine-tuned with domain tags serves as a very competitive baseline for multilingual domain adaptation. On the other hand, domain adaptation with adapters offers modularity, and allows incrementally adapting to new domains without retraining the full model. Future research directions could explore multi-task training combining parallel and monolingual in-domain data in other ways to alleviate the need for back-translation. 

Our work is the first attempt to combine domain adapters and language adapters for a generation task (NMT). Although such combinations have shown to be successful for NLU tasks, obtaining good representations for generating unseen target languages proves to be a difficult problem. We believe a fine-grained study of where to use language or domain-specific capacity could lead to better cross-lingual domain transfer in future. 
Finally, we provide supplementary material to facilitate reproducibility.\footnote{\url{https://tinyurl.com/r66stbxj}} 
\section*{Acknowledgements}

We would like to thank Laurent Besacier, 
Hady Elsahar, Matthias Gallé, Germán Kruszewski, Ahmet Ustun and all of the NAVER Labs Europe team for useful discussions. Asa Cooper Stickland was supported in part by the EPSRC Centre for Doctoral Training in
Data Science, funded by the UK Engineering and Physical Sciences Research Council (grant EP/L016427/1) and the University of Edinburgh.

\bibliography{anthology,custom,main}
\bibliographystyle{acl_natbib}
\newpage
\newpage
\clearpage
\appendix

\section{Data and Hyper-parameters}
\label{sec:appendix}
For \textbf{bilingual} domain adaptation we use a Transformer Base \cite{vaswani2017attention} model trained for 12 epochs on German to English WMT20 data (47M parallel lines), with a joint BPE \cite{sennrich2016neural} vocabulary of size 24k with inline casing \cite{berard2019} (i.e.\ wordpieces are put in lowercase with a special token indicating their case.).
For bilingual domain adaption we use the same datasets as \citet{aharoni2020unsupervised}, namely parallel text in German and English from five diverse domains: Koran, Medical, IT, Law and Subtitles.

For multilingual settings we use the following hyper-parameters. We share embeddings between encoder and decoder. We use the Adam optimizer \cite{kingma2014adam} with an inverse square root learning rate schedule for pre-training, and a fixed learning rate schedule for training adapters. We speed up training with 16 bit floating point arithmetic. We use label smoothing 0.1 and dropout 0.1. We train for either 20 epochs or 1 million updates, whichever corresponds to the smallest number of training updates. We use early stopping, checking performance after each epoch or every 100,000 training steps, and use average validation negative-log-likelihood on all of the training data (but not out-of-domain language data) as our criteria for choosing the best model. We otherwise use default Fairseq \cite{ott-etal-2019-fairseq} parameters. We train all models on a single Nvidia V100 GPU, and training takes between 8 and 36 hours depending on dataset size.  

In order to create validation and test splits that had no overlap with training data in any language, we first set aside a number of English sentences. Then we aligned all language pairs to these sentences, i.e.\ the German to French test set is composed of German and French sentences that share the same English sentence. Finally we remove all sentences in any language from the train splits of all parallel data if those sentences are aligned with any English sentences in the subset we set aside for validation/test splits. Both validation sets and test sets contain around 2000 examples for every domain and language-pair.
\section{MAD-X Style Stacking}
\label{sec:mad}
\citet{pfeiffer-etal-2020-mad} use the following stacking formulation,
\begin{equation}
    \textrm{LA}(\textbf{h}_l, \textbf{r}_l) = \textrm{FFN}_{\textrm{lg}}(\textbf{h}_l) + \textbf{r}_l.
\end{equation}
The residual connection $\textbf{r}_l$ is the output of the Transformer's feed-forward layer whereas $\textbf{h}_l$ is the output of the subsequent layer normalisation. When stacking domain and language adapters the layer output is given by applying the model's pre-trained layer norm $\textrm{LN}_{\textrm{pre}}$, 
\begin{equation}
\label{eq:pfieff}
\textbf{h}_{l,\textrm{out}} = \textrm{LN}_{\textrm{pre}}(\textrm{FFN}_{\textrm{dom}}(\textrm{LA}(\textbf{h}_l, \textbf{r}_l)) + \textbf{r}_l)
\end{equation}
and using the output of the Transformer's feed-forward layer as a residual instead of the language adapter output. We refer to this as `\textbf{MAD-X}' style after \citet{pfeiffer-etal-2020-mad}. This leaves the layer output `closer' to the pre-trained model, with the same layer-norm and residual connection, contrary to Eq.~\ref{eq:bapna} which has a newly initialised layer-norm and a residual connection. For all models without any stacking we obtain layer output as in Eq.~\ref{eq:pfieff} but replace $\textrm{LA}(\cdot)$ with the identity operation.

\section{Additional Results for Bilingual Domain Adaptation}
Before studying multilingual domain adaptation, we validate some of our ideas on a simpler, \emph{bilingual} German $\to$ English domain adaptation setting. Table~\ref{tab:bi0} reports the results of this experiment. First, we note that \emph{encoder-only} adapters perform similarly to \emph{encoder \& decoder} adapters, while \emph{decoder-only} adapters perform worse. 

Moreover, adding adapters to only the last three layers of the encoder almost matches the performance of adapting every layer, while adding adapters to the first three layers decreases performance.  We believe this is because the last encoder layer directly influences every layer of the decoder through cross-attention.

Table~\ref{tab:bi} presents results of bilingual domain adaption with smaller adapter bottleneck dimension. The same trends emerge: encoder-only adapters perform better, and the last three layers of the encoder are better than the first three. The last three encoder layers also perform better than the first three for a multilingual model, see Table~\ref{tab:frdecs-medical2} models \modelref{f3} and \modelref{l3}. Interestingly the multilingual last three encoder layer DA model is roughly halfway between encoder-only and decoder-only on Out$\to$in and In$\to$out\ performance, suggesting it might be a useful compromise between the two. 
\label{sec:bil2}
\begin{figure*}
    \centering
    \includegraphics[width=\linewidth]{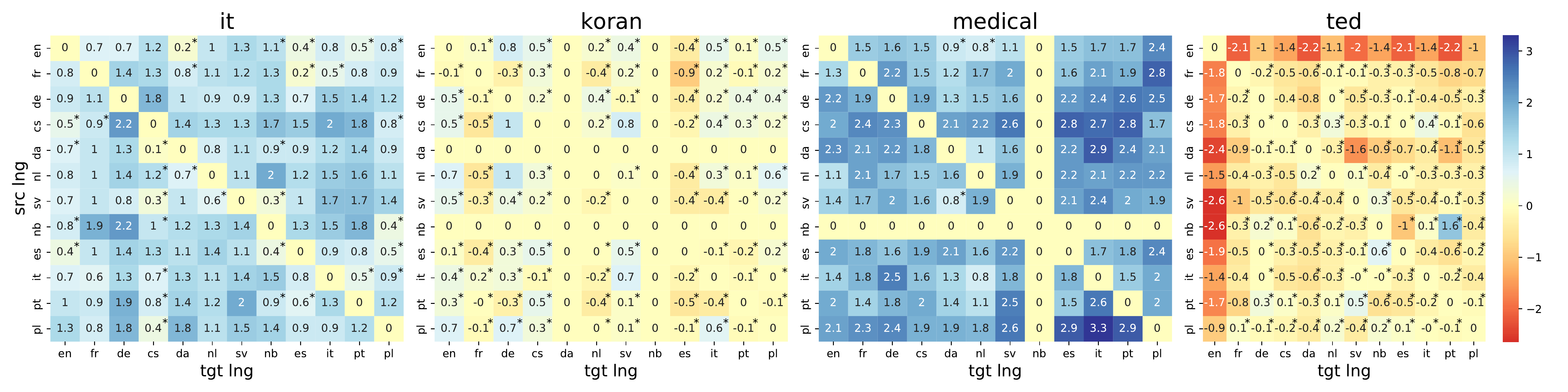}
    \caption{Difference in BLEU score for each domain between the model trained with adapters \modelref{multiLA_DA} and model trained with domain tags \modelref{domain_tags_all}, for the multilingual multi-domain models. A positive number corresponds to the case where model \modelref{domain_tags_all} has higher score than the model \modelref{multiLA_DA}. The "*" indicates the cases when the difference is \textit{not} statistically significant. }
    \label{fig:mmd_st}
\end{figure*}
\section{Additional Results for Cross-lingual Transfer}
\label{sec:add}
\begin{table*}[]
    \centering
    \begin{tabular}{|p{3.5cm}|p{3.5cm}|p{3.5cm}|p{3.5cm}|}\hline
        source (fr) & ref (pt)  & \modelref{vanillaDA} & \modelref{EncDecDA}\\\hline
         \small{La durée du traitement dépend de la nature et de la sévérité de l’ infection et de la réponse observée.}&\small{A duração do tratamento depende da natureza e da gravidade da infecção e da resposta verificada.}&\small{The duration of treatment depends on the nature and severity of the infection and on the response observed.}&\small{A duração do tratamento depende da nature e severidade da infecção e da resposta observed.}\\\hline
         \small{Insuman Comb 50 40 UI/ ml suspension injectable en flacon} &\small{Insuman Comb 50 40 UI/ ml, Suspensão injectável num frasco para injectáveis} &\small{Insuman Comb 50 40 IU/ ml suspension injectable en flacon} &\small{Insuman Comb 50 40 IU/ ml suspension for injection in vial}
\\\hline
         \small{A quoi ressemble TAXOTERE et contenu de l’ emballage extérieur TAXOTERE 80 mg, solution à diluer pour perfusion est une solution visqueuse, limpide, jaune à jaune marron.}&\small{Qual o aspecto de TAXOTERE e contéudo da embalagem TAXOTERE 80 - mg concentrado para solução para perfusão é uma solução viscosa transparente amarela ou amarela- acastanhada.}&\small{What TAXOTERE looks like and contents of the pack TAXOTERE 80 mg concentrate para solution for infusion is a solution visqueuse, limpida, de jaune à marron.}& \small{TAXOTERE 80 mg, Diluted for Solution for Infusion é uma solution viscous, limpa, yellow to marrom.}\\\hline

    \end{tabular}
    \caption{Some examples of translations generated by straight-forward adapter training settings, in this case from a known source language, \textit{fr} into a target language unseen during domain adaptation, \textit{pt}, and for the medical domain.}
    \label{tab:ex_offtarget}
\end{table*}

\paragraph{Does language diversity increase transfer?} Table~\ref{tab:comp} compares models trained on a mix of language families (fr, de, cs, en) and mostly romance languages (fr, it, es, en) to test whether diversity of languages in our in-domain training set improved transfer. Positive numbers in this table indicate diversity of training languages improves performance. Diversity helps for translating out-of-domain languages into in-domain. We have unclear results for when both source and target are out-of-domain; it seems when using back-translation (BT), i.e.\ when all languages have been seen (albeit with artificial English parallel data) diversity helps, but without BT it mostly hurts performance. We speculate that training on mostly romance languages means the domain adapter encodes less `language information', but leave further exploration to future work.

\begin{table}
\centering
\begin{tabular}{ccc}
\toprule
Model & Out$\to$ $\{$en,fr$\}$ & Out$\to$Out \\
\midrule
\textbf{Koran} & \\
LA + Dec. DA & 0.6 & -0.9 \\
LA + Dec. DA & 0.3 & -0.3 \\
Unfr. LA + Dec. DA & 0.1 & -0.4 \\
LA + Enc \& Dec. DA & 1.3 & -1.3 \\
\midrule
\textbf{Koran + BT} & \\
LA + Dec. DA & 0.4 & 0.2 \\
LA + Dec. DA & 1.9 & 0.9 \\
Unfr. LA + Dec. DA & 0.3 & 0 \\
LA + Enc \& Dec. DA & 0.7 & 0.3 \\
\midrule
\textbf{Medical} & \\
LA + Dec. DA & 0.8 & -2.4 \\
LA + Dec. DA & 3.2 & 0.2 \\
Unfr. LA + Dec. DA & 0.2 & 0.1 \\
LA + Enc \& Dec. DA & 3.2 & -1.1 \\
\midrule
\textbf{Medical + BT} & \\
LA + Dec. DA & 0.4 & 2.9 \\
LA + Dec. DA & 1.3 & 1.6 \\
Unfr. LA + Dec. DA & 0.5 & 3.1 \\
LA + Enc \& Dec. DA & 0.5 & 1.5 \\ 
\end{tabular} 
\caption{\label{tab:comp}
Difference in average BLEU score between models trained on a diverse subset of languages and models trained on mostly romance languages. Data source is noted in bold. Refer to the main paper for model definitions. Out$\to$ $\{$en,fr$\}$ corresponds to translation from an out-of-domain source language into $\{$en,fr$\}$. `Out$\to$Out' is the average score when both the source and target  language are unseen during domain adaptation (choosing languages unseen by either subset).
}
\end{table}

\paragraph{Additional results and metrics} We present additional results for the setting discussed in Section~\ref{sec:offtgt} of the main paper in Table~\ref{tab:frdecs-koran3} (Koran domain), Table~\ref{tab:frdecs-koran2} (Koran results for the romance language subset), and Table~\ref{tab:frdecs-medical2} (additional Medical results). We use the chrF metric as discussed in the main paper, and find the conclusions based on BLEU score are unchanged. For the Koran domain, we see similar trends with decoder-only domain adapters (DA) performing best on out-of-domain source to in-domain target languages, and vice versa for encoder-only DA. Additionally we see as before that combining BT with decoder-only DA works the best, and achieves the highest overall performance. We report on-target (correct language) percentage for all medical domain models in Table~\ref{tab:frdecs-medical2-on-target}.

We briefly experiment with denoising objectives, where we simply copy target data in out-of-domain languages to the source side (and optionally add `noise' to the source side, e.g.\ swap tokens or mask tokens \cite{lewis2019bart}). Although we got reasonable improvements (models \modelref{monoenc} and \modelref{monodec}) for out-of-domain target languages, we were mostly unable to improve over the pre-trained ParaCrawl LA, and so concentrate on back-translation. 

We experiment with a setting where we jointly train on all language directions for IT, Koran and TED Talks domains and a subset of languages for Medical, and similarly with only a subset of Koran (models \modelref{multidec}, \modelref{multienc} etc.). These models stack language and domain adapters. Such models don't require any pre-trained LA, and improve out-of-domain performance and decrease off-target translation compared to freezing ParaCrawl LA and training DA. However these scores are still worse than simply using pre-trained `domain-agnostic' ParaCrawl LA \modelref{ParaCrawlLA}.

\begin{table*}[ht!]
\centering
\begin{tabular}{clccccc}
\toprule
ID & Model & IT & Koran & Medical & TED & Params (M)\\
\midrule
\modelref{base} & Base (En-centric) & .456 & .300 & .488 & .456 & N/A \\
\midrule
\modelref{finetune_all_domains} & Finetuned & .623 & .394 & .625 & .513 & 79 \\
\modelref{domain_tags_all} & Finetuned + domain tags & \textbf{.645} & \textbf{.433} & \textbf{.646} & .517 & 79 \\
\midrule
\modelref{singleLA} & Single adapter per layer ($d=1024$) & .612 & .382 & .619 & .512 & 12.6 \\
\modelref{multiLA} & LA ($d=1365$) & .632 & .408 & .630 & .517 & 202 \\
\modelref{multiLA_2048} & LA ($d=2048$) & .634 & .411 & .631 & .517 & 303  \\
\modelref{multiLA_decDA}  & LA + dec. DA ($d=1024$) & .633 & .412 & .629 & .518 & 177 \\
\modelref{multiLA_encDA} & LA + enc. DA ($d=1024$) & .634 & .426 & .631 & .522 & 177 \\
\modelref{multiLA_DA} & LA + enc \& dec. DA ($d=1024$) & .636 & .429 & .632 & \textbf{.523} & 202\\
\end{tabular} 
\caption{\label{tab:multi_chrf}
chrF scores of various multilingual multi-domain adaptation strategies, i.e. training on all language directions from the 12 languages and all domains.}
\end{table*}

\begin{table*}[ht!]
\centering
\begin{tabular}{clccccc}
\toprule
ID & Model & All & In$\to$in & Out$\to$in & In$\to$out& Out$\to$out \\
\midrule
& \multicolumn{1}{c}{\textit{Oracles}} \\
\midrule
\modelref{finetune_all_langs} & Finetune (all langs) & .635 & .631 & .638 & .635 & .635 \\
\modelref{domain_tags_all} & FT (all langs \& domains) + domain tags & .646 & .641 & .648 & .646 & .646 \\
\midrule
& \multicolumn{1}{c}{\textit{Baselines}} \\
\midrule
\modelref{base} & Base (En-centric) & .488 & .500 & .497 & .493 & .475 \\
\modelref{ParaCrawlLA}  & \modelref{base} + ParaCrawl LA & .537 & .532 & .540 & .538 & .535 \\
\midrule
& \multicolumn{1}{c}{\textit{Straightforward Methods}} \\
\midrule
\modelref{vanillaDA} & \modelref{base} + Domain adapters only & .400 & .635 & .575 & .295 & .287 \\
\modelref{EncDecDA} & Freeze LA + enc. \& dec. DA & .468 & .631 & .566 & .401 & .400 \\
\modelref{TagsCross_no_paracrawl} & FT (all domains) + dom. tags & .277 & \bf{.650} & .236 & .283 & .193 \\
\midrule
& \multicolumn{1}{c}{\textit{Improving Off-target Translation}} \\
\midrule
\modelref{TagsCross} & \modelref{TagsCross_no_paracrawl} + ParaCrawl & .570 & .622 & .601 & .556 & .544 \\
\labelledmodelcounter{unfreezeLA} & Unfreeze LA & .551 & .639 & .574 & .514 & .535 \\
\labelledmodelcounter{unfreezeLABT} & \modelref{unfreezeLA} + BT & .571 & .636 & .556 & .594 & .548 \\
\midrule
\modelref{freezeLA_decDA} & Freeze LA + dec. DA & .492 & .613 & \bf{.605} & .432 & .421 \\
\modelref{freezeLA_decDA_BT} & \modelref{freezeLA_decDA} + BT & \bf{.586} & .608 & .590 & .584 & \bf{.577} \\
\labelledmodelcounter{freezeLA_decDA_DADrop_BT} & \modelref{freezeLA_decDA} + BT + DADrop & .584 & .604 & .587 & .583 & .576 \\
\modelref{freezeLA_encDA} & Freeze LA + enc. DA & .518 & .623 & .548 & .506 & .477 \\
\modelref{freezeLA_encDA_BT} & \modelref{freezeLA_encDA} + BT & .548 & .620 & .567 & .574 & .498 \\
\labelledmodelcounter{freezeLA_encDA_DADrop_BT} & \modelref{freezeLA_encDA} + BT + DADrop & .561 & .614 & .570 & .579 & .527 \\
\labelledmodelcounter{f3} & Freeze LA + enc. first 3 layers DA & .440 & .618 & .525 & .379 & .373 \\
\labelledmodelcounter{l3} & Freeze LA + enc. last 3 layers DA & .512 & .622 & .576 & .472 & .465 \\ 
\midrule
\modelref{EncDecDA_DADrop} & \modelref{EncDecDA} + DADrop & .490 & .621 & .567 & .447 & .429 \\
\labelledmodelcounter{EncDecDA_BT_chrf} & \modelref{EncDecDA} + BT & .559 & .626 & .581 & .582 & .511 \\
\modelref{EncDecDA_DADrop_BT} & \modelref{EncDecDA} + BT + DADrop & .569 & .619 & \bf{.583} & .587 & .534 \\ 
\labelledmodelcounter{EncDecDA_BT_pfeiffer} & \modelref{EncDecDA} + BT + MAD-X style & .540 & .625 & .575 & .561 & .477 \\
\midrule
\modelref{unfreezeLA_decDA} & Unfreeze LA + dec. DA & .221 & .641 & .573 & .010 & .007 \\
\modelref{unfreezeLA_decDA_DADrop} & \modelref{unfreezeLA_decDA} + DADrop & .528 & .639 & .577 & .452 & .514 \\
\modelref{unfreezeLA_decDA_DADrop_BT} & \modelref{unfreezeLA_decDA} + DADrop + BT & .573 & .636 & .559 & .595 & .550 \\
\end{tabular} 
\caption{\label{tab:frdecs-medical2}
chrF score of various models trained on the $\{$en, fr, de, cs$\}$ subset of the Medical domain. Some models are also included in the main paper.}
\end{table*}

\begin{table*}[ht!]
\centering
\begin{tabular}{clccccc}
\toprule
ID & Model & All & In$\to$in & Out$\to$in & In$\to$out& Out$\to$out \\
\midrule
& \multicolumn{1}{c}{\textit{Oracles}} \\
\midrule
\modelref{finetune_all_langs} & Finetune (all langs) & 95\% & 95\% & 95\% & 95\% & 95\% \\
\modelref{domain_tags_all} & FT (all langs \& domains) + domain tags & 95\% & 95\% & 96\% & 95\% & 95\% \\
\midrule
& \multicolumn{1}{c}{\textit{Baselines}} \\
\midrule
\modelref{base} & Base (En-centric) & 91\% & 93\% & 92\% & 91\% & 89\% \\
\modelref{ParaCrawlLA}  & \modelref{base} + ParaCrawl LA & 95\% & 96\% & 96\% & 95\% & 95\% \\
\midrule
& \multicolumn{1}{c}{\textit{Straightforward Methods}} \\
\midrule
\modelref{vanillaDA} & \modelref{base} + Domain adapters only & 40\% & 95\% & 93\% & 7\% & 11\% \\
\modelref{EncDecDA} & Freeze LA + enc. \& dec. DA & 81\% & 95\% & 93\% & 71\% & 76\% \\
\modelref{TagsCross_no_paracrawl} & FT (all domains) + dom. tags & 25\% & 96\% & 55\% & 1\% & 2\% \\
\midrule
& \multicolumn{1}{c}{\textit{Improving Off-target Translation}} \\
\midrule
\modelref{TagsCross} & \modelref{TagsCross_no_paracrawl} + ParaCrawl & 93\% & 95\% & 93\% & 93\% & 92\% \\
\labelledmodelcounter{unfreezeLA} & Unfreeze LA & 94\% & 95\% & 95\% & 93\% & 95\% \\
\labelledmodelcounter{unfreezeLABT} & \modelref{unfreezeLA} + BT & 94\% & 96\% & 95\% & 95\% & 94\% \\
\midrule
\modelref{freezeLA_decDA} & Freeze LA + dec. DA & 84\% & 95\% & 95\% & 77\% & 77\% \\
\modelref{freezeLA_decDA_BT} & \modelref{freezeLA_decDA} + BT & 94\% & 95\% & 95\% & 94\% & 94\% \\
\labelledmodelcounter{freezeLA_decDA_DADrop_BT} & \modelref{freezeLA_decDA} + BT + DADrop & 94\% & 95\% & 95\% & 94\% & 94\% \\
\modelref{freezeLA_encDA} & Freeze LA + enc. DA & 90\% & 95\% & 93\% & 89\% & 88\% \\
\modelref{freezeLA_encDA_BT} & \modelref{freezeLA_encDA} + BT & 90\% & 96\% & 94\% & 94\% & 83\% \\
\labelledmodelcounter{freezeLA_encDA_DADrop_BT} & \modelref{freezeLA_encDA} + BT + DADrop & 93\% & 95\% & 95\% & 94\% & 91\% \\
\labelledmodelcounter{f3} & Freeze LA + enc. first 3 layers DA & 59\% & 95\% & 93\% & 35\% & 43\% \\
\labelledmodelcounter{l3} & Freeze LA + enc. last 3 layers DA & 88\% & 95\% & 94\% & 83\% & 86\% \\
\midrule
\modelref{EncDecDA_DADrop} & \modelref{EncDecDA} + DADrop & 86\% & 95\% & 93\% & 82\% & 82\% \\
\modelref{EncDecDA_BT} & \modelref{EncDecDA} + BT & 91\% & 95\% & 95\% & 94\% & 85\% \\
\modelref{EncDecDA_DADrop_BT} & \modelref{EncDecDA} + BT + DADrop & 93\% & 95\% & 95\% & 94\% & 91\% \\
\labelledmodelcounter{EncDecDA_BT_pfeiffer} & \modelref{EncDecDA} + BT + MAD-X style & 89\% & 95\% & 95\% & 92\% & 80\% \\
\midrule
\modelref{unfreezeLA_decDA} & Unfreeze LA + dec. DA & 36\% & 96\% & 95\% & 1\% & 2\% \\
\modelref{unfreezeLA_decDA_DADrop} & \modelref{unfreezeLA_decDA} + DADrop & 90\% & 95\% & 95\% & 82\% & 92\% \\
\modelref{unfreezeLA_decDA_DADrop_BT} & \modelref{unfreezeLA_decDA} + DADrop + BT & 94\% & 95\% & 95\% & 94\% & 94\% \\
\end{tabular} 
\caption{\label{tab:frdecs-medical2-on-target}
On-target translation percentages of various models trained on the $\{$en, fr, de, cs$\}$ subset of the Medical domain.}
\end{table*}

\begin{table*}[ht!]
\centering
\begin{tabular}{clccccc}
\toprule
ID & Model & All & In$\to$in & Out$\to$in & In$\to$out& Out$\to$out \\
\midrule
& \multicolumn{1}{c}{\textit{Oracles}} \\
\midrule
\modelref{finetune_all_langs} & Finetune (all langs) & .461 & .437 & .427 & .487 & .477 \\
\modelref{domain_tags_all} & FT (all langs \& domains) + domain tags & .433 & .423 & .409 & .455 & .438 \\
\midrule
& \multicolumn{1}{c}{\textit{Baselines}} \\
\midrule
\modelref{base} & Base (En-centric) & .300 & .307 & .299 & .306 & .294 \\
\modelref{ParaCrawlLA} & \modelref{base} + ParaCrawl LA & .334 & .330 & .328 & .340 & .335 \\
\midrule
& \multicolumn{1}{c}{\textit{Straightforward Methods}} \\
\midrule
\modelref{vanillaDA} & \modelref{base} + Domain adapters only & .246 & .451 & .349 & .163 & .150 \\
\modelref{EncDecDA} & Freeze LA + enc. \& dec. DA & .165 & .449 & .137 & .144 & .089 \\
\midrule
& \multicolumn{1}{c}{\textit{Improving Off-target Translation}} \\
\midrule
\modelref{TagsCross_no_paracrawl} & FT (all dom.) + dom. tags & .166 & .436 & .162 & .143 & .081 \\
\modelref{TagsCross} & FT (all dom. + ParaCrawl) + dom. tags & .359 & .410 & .375 & .351 & .332 \\
\modelref{unfreezeLA} & Unfreeze LA & .352 & .454 & .351 & .322 & .335 \\
\midrule
\modelref{freezeLA_decDA} & Freeze LA + dec. DA & .304 & .404 & \bf{.385} & .249 & .244 \\
\labelledmodelcounter{monodec} & \modelref{freezeLA_decDA} + Mono data & .355 & .390 & .373 & .342 & .336 \\
\modelref{freezeLA_decDA_BT} & \modelref{freezeLA_decDA} + BT & .381 & .399 & .371 & .387 & .375 \\
\modelref{freezeLA_decDA_DADrop_BT} & \modelref{freezeLA_decDA} + BT + DADrop & \bf{.382} & .402 & .373 & .388 & \bf{.376} \\
\modelref{freezeLA_encDA} & Freeze LA + enc. DA & .319 & .438 & .328 & .315 & .266 \\
\labelledmodelcounter{monoenc} & \modelref{freezeLA_encDA} + Mono data & .347 & .410 & .338 & .353 & .324 \\ 
\modelref{freezeLA_encDA_BT} & \modelref{freezeLA_encDA} + BT & .365 & .432 & .353 & .385 & .330 \\
\modelref{freezeLA_encDA_DADrop_BT} & \modelref{freezeLA_encDA} + BT + DADrop & .368 & .425 & .354 & \bf{.394} & .336 \\
\midrule
\modelref{EncDecDA_BT} & \modelref{EncDecDA} + BT & .374 & .434 & .366 & \bf{.394} & .341 \\
\modelref{EncDecDA_DADrop_BT} & \modelref{EncDecDA} + BT + DADrop & .381 & .436 & .369 & .406 & .349 \\
\midrule
\modelref{unfreezeLA_decDA} & Unfreeze LA + dec. DA & .224 & .457 & .351 & .088 & .138 \\
\modelref{unfreezeLA_decDA_DADrop} & \modelref{unfreezeLA_decDA} + DADrop & .339 & \bf{.458} & .354 & .288 & .320 \\
\midrule
\labelledmodelcounter{multidec} & Multi-domain dec. DA & .326 & .403 & .360 & .304 & .285 \\
\labelledmodelcounter{multienc} & Multi-domain enc. DA & .337 & .412 & .360 & .327 & .297 \\
\labelledmodelcounter{multiencdec} & Multi-domain enc. \& dec. DA & .327 & .417 & .369 & .302 & .279 \\ 

\end{tabular} 
\caption{\label{tab:frdecs-koran3}
chrF score of various models trained on the $\{$en, fr, de, cs$\}$ subset of the Koran domain. LA = language adapters, DA = domain adapters. `Out$\to$in' is the average score when translating from an out-of-domain source language into $\{$en, fr, de, cs$\}$. `In$\to$out' corresponds to when the out-of-domain language is the target language. `In$\to$in' refers to average score when source and target are in the set $\{$en, fr, de, cs$\}$. `Out$\to$Out' is the average score when both the source and target  language are unseen during domain adaptation. `Mono data' refers to adding copied monolingual data for out-of-domain languages, and additionally multiparallel ParaCrawl data in small amounts.}
\end{table*}

\begin{table*}[ht!]
\centering
\begin{tabular}{clccccc}
\toprule
ID & Model & All & In$\to$in & Out$\to$in & In$\to$out& Out$\to$out \\
\midrule
\modelref{EncDecDA}  & Freeze LA + enc. \& dec. DA & .309 & .491 & .362 & .267 & .229 \\
\modelref{EncDecDA_DADrop} & \modelref{EncDecDA} + DADrop & .311 & .490 & .360 & .270 & .233 \\
\modelref{unfreezeLA} & Unfreeze LA & .357 & .515 & .395 & .303 & .307 \\
\modelref{unfreezeLABT} & \modelref{unfreezeLA} + BT & .372 & .529 & .364 & .370 & .319 \\
\midrule
\modelref{freezeLA_decDA} & Freeze LA + dec. DA & .332 & .463 & \bf{.418} & .268 & .262 \\
\modelref{freezeLA_decDA_BT} & \modelref{freezeLA_decDA} + BT & \bf{.382} & .460 & .408 & .362 & \bf{.345} \\ 
\modelref{freezeLA_encDA} & Freeze LA + enc. DA & .320 & .491 & .345 & .296 & .249 \\
\modelref{freezeLA_encDA_BT} & \modelref{freezeLA_encDA} + BT & .359 & .492 & .374 & .354 & .298 \\
\midrule
\modelref{unfreezeLA_decDA} & Unfreeze LA + dec. DA & .322 & .519 & .399 & .216 & .267 \\
\modelref{unfreezeLA_decDA_DADrop} & \modelref{unfreezeLA_decDA} + DADrop & .353 & .515 & .399 & .291 & .303 \\
\modelref{unfreezeLA_decDA_DADrop_BT} & \modelref{unfreezeLA_decDA} + DADrop + BT & .377 & \bf{.522} & .372 & \bf{.373} & .327 \\
\modelref{EncDecDA_BT} & \modelref{EncDecDA} + BT & .368 & .490 & .388 & .363 & .308 \\
\modelref{EncDecDA_DADrop_BT} & \modelref{EncDecDA_BT} + DADrop & .373 & .494 & .389 & .369 & .314 \\
\end{tabular} 
\caption{\label{tab:frdecs-koran2}
chrF score of various models trained on the mostly romance language $\{$en, fr, it, es$\}$ subset of the Koran domain. LA = language adapters, DA = domain adapters. `Out$\to$in' is the average score when translating from an out-of-domain source language into $\{$en, fr, it, es$\}$. `In$\to$out' corresponds to when the out-of-domain language is the target language. `In$\to$in' refers to average score when source and target are in the set $\{$en, fr, it, es$\}$. `Out$\to$Out' is the average score when both the source and target  language are unseen during domain adaptation.}
\end{table*}

\begin{table*}[ht!]
\centering
\begin{tabular}{clccccc}
\toprule
ID & Model & IT & Koran & Medical & Subtitles & Law \\

\midrule
\modelcounter & No fine-tuning & 35.3 & 14.8 & 38.1 & 26.8 & 42.4 \\
\modelcounter & Fine-tuned & 43.8 & 22.7 & 53 & 30.9 & 57.9 \\
\midrule
\labelledmodelcounter{enc_dec_bi} & Enc. + dec. adapters ($d=1024$) & 42.9 & 21.8 & 51.7 & 30.5 & 56 \\
\modelcounter & \modelref{enc_dec_bi} + MAD-X style  & 40.6 & 19.3 & 48.8 & 29.8 & 54.3 \\
\modelcounter & Dec. adapters ($d=2048$) & 42.1 & 19.8 & 50.5 & 29.7 & 55.1 \\
\labelledmodelcounter{enc_bi} & Enc. adapters ($d=2048$)  & 42.4 & 21.5 & 51.9 & 30.1 & 56.1 \\
\modelcounter & Last 3 encoder layers only ($d=4096$) & 42.9 & 21.1 & 52.1 & 30.1 & 56 \\
\modelcounter & First 3 encoder layers only ($d=4096$) & 42.2 & 20 & 50.1 & 28.5 & 54.9 \\
\end{tabular} 
\caption{\label{tab:bi0}
BLEU scores of various domain adaptation strategies for a German $\to$ English bilingual model. $(d=N)$ refers to adapters with a bottleneck dimension of size $N$.}
\end{table*}

\begin{table*}[ht!]
\centering
\begin{tabular}{clccccc}
\toprule
ID & Model & IT & Medical & Koran & Subtitles & Law \\
\midrule
\modelcounter & No fine-tuning & 35.3 & 14.8 & 38.1 & 26.8 & 42.4 \\
\modelcounter & Finetuned & 43.8 & 22.7 & 53 & 30.9 & 57.9 \\
\modelcounter & Enc. + dec. adapters ($d$=64) & 40 & 18.7 & 47.3 & 29.4 & 51.5 \\
\modelcounter & Dec. adapters ($d$=128) & 39 & 17.5 & 46 & 28.8 & 50.6 \\
\modelcounter & Enc. adapters ($d$=128) & 40 & 18.9 & 47.3 & 29.2 & 51.5 \\
\modelcounter & Last 3 encoder layers only ($d$=256) & 40 & 19 & 47.3 & 29 & 51.1 \\
\modelcounter & First 3 encoder layers only ($d$=256) & 39.5 & 18 & 46 & 28.8 & 49.5 \\
\end{tabular} 
\caption{\label{tab:bi}
BLEU scores of various domain adaptation strategies for a German $\to$ English bilingual model. $(d=N)$ refers to adapters with a bottleneck dimension of size $N$.}
\end{table*}

\end{document}